\documentclass[10pt,twocolumn,letterpaper]{article}

\usepackage[pagenumbers]{iccv}              

\definecolor{iccvblue}{rgb}{0.21,0.49,0.74}

\usepackage{amssymb}
\usepackage{amsmath}
\usepackage{pifont}
\usepackage{bm}
\usepackage{booktabs}
\usepackage{mathtools}
\usepackage{multirow}

\usepackage[numbers]{natbib}
\usepackage[resetlabels]{multibib}
\newcites{supp}{Supplementary References}
\usepackage[pagebackref,breaklinks,colorlinks,citecolor=iccvblue,allcolors=iccvblue]{hyperref}
\usepackage{etoolbox}
\makeatletter
\patchcmd\Hy@backout{\@auxout}{\@mainaux}{}{\fail}
\patchcmd\Hy@backout{\@auxout}{\@mainaux}{}{\fail}
\makeatother

\newcommand{\myparagraph}[1]{\vspace{2mm}\noindent\textbf{#1}}

\usepackage{balance}
\usepackage{xcolor}
\usepackage{color,colortbl}

\definecolor{col_nose}{RGB}{181,228,140}
\definecolor{col_mouth}{RGB}{255,183,3}
\definecolor{col_forehead}{RGB}{189,178,255}
\definecolor{col_cheek}{RGB}{144,224,239}
\definecolor{col_table}{RGB}{175,227,246}

\definecolor{mycolor1}{rgb}{0.85000,0.32500,0.09800}
\definecolor{mycolor2}{rgb}{0.92900,0.69400,0.12500}
\definecolor{mycolor3}{rgb}{0.49400,0.18400,0.55600}
\definecolor{mycolor4}{rgb}{0.87843,0.76471,0.98824}
\definecolor{mycolor5}{rgb}{0.46600,0.67400,0.18800}
\definecolor{mycolor6}{rgb}{0.30100,0.74500,0.93300}
\definecolor{mycolor7}{rgb}{0.00000,0.44700,0.74100}

\definecolor{best_two}{RGB}{72,149,239}
\definecolor{best}{RGB}{179,11,0}

\usepackage[lined,ruled,linesnumbered]{algorithm2e}

\usepackage{threeparttable}

\title{BFSM: 3D Bidirectional Face-Skull Morphable Model}

\author{Zidu Wang$^{1,2}$, Meng Xu$^{3}$, Miao Xu$^{1,2,4}$, Hengyuan Ma$^{3}$,\\ Jiankuo Zhao$^{1,2}$, Xutao Li$^{1,2}$, Xiangyu Zhu$^{1,2}$\footnotemark[1], Zhen Lei$^{1,2,4}$\\
    $^{1}$State Key Laboratory of Multimodal Artificial Intelligence Systems,\\ Institute of Automation, Chinese Academy of Sciences\\
    $^{2}$School of Artificial Intelligence, University of Chinese Academy of Sciences\\
    $^{3}$ Plastic Surgery Hospital, Chinese Academy of Medical Sciences and Peking Union Medical College \\  
    $^{4}$ Centre for Artificial Intelligence and Robotics, Hong Kong Institute of Science \& Innovation,\\ Chinese Academy of Sciences\\
    {\tt\small\{wangzidu0705, xum8701\}@gmail.com, {\tt\small mahy@psh.pumc.edu.cn},}\\ {\tt\small\{xumiao2021, zhaojiankuo2024, lixutao2025, xiangyu.zhu, zhen.lei\}@ia.ac.cn}
}

\begin{document}

\maketitle

{\renewcommand{\thefootnote}{\fnsymbol{footnote}}
    \footnotetext[1]{Corresponding author.}}

\begin{abstract}{

Building a joint face–skull morphable model holds great potential for applications such as remote diagnostics, surgical planning, medical education, and physically based facial simulation. However, realizing this vision is constrained by the scarcity of paired face–skull data, insufficient registration accuracy, and limited exploration of reconstruction and clinical applications. Moreover, individuals with craniofacial deformities are often overlooked, resulting in underrepresentation and limited inclusivity. To address these challenges, we first construct a dataset comprising over 200 samples, including both normal cases and rare craniofacial conditions. Each case contains a CT-based skull, a CT-based face, and a high-fidelity textured face scan. Secondly, we propose a novel dense ray matching registration method that ensures topological consistency across face, skull, and their tissue correspondences. Based on this, we introduce the 3D Bidirectional Face–Skull Morphable Model (BFSM), which enables shape inference between the face and skull through a shared coefficient space, while also modeling tissue thickness variation to support one-to-many facial reconstructions from the same skull, reflecting individual changes such as fat over time. Finally, we demonstrate the potential of BFSM in medical applications, including 3D face–skull reconstruction from a single image and surgical planning prediction. Extensive experiments confirm the robustness and accuracy of our method. BFSM is available at \href{https://github.com/wang-zidu/BFSM}{https://github.com/wang-zidu/BFSM}.

}
\end{abstract}

\begin{figure}[t]
\centering
\includegraphics[width=1.0\columnwidth]{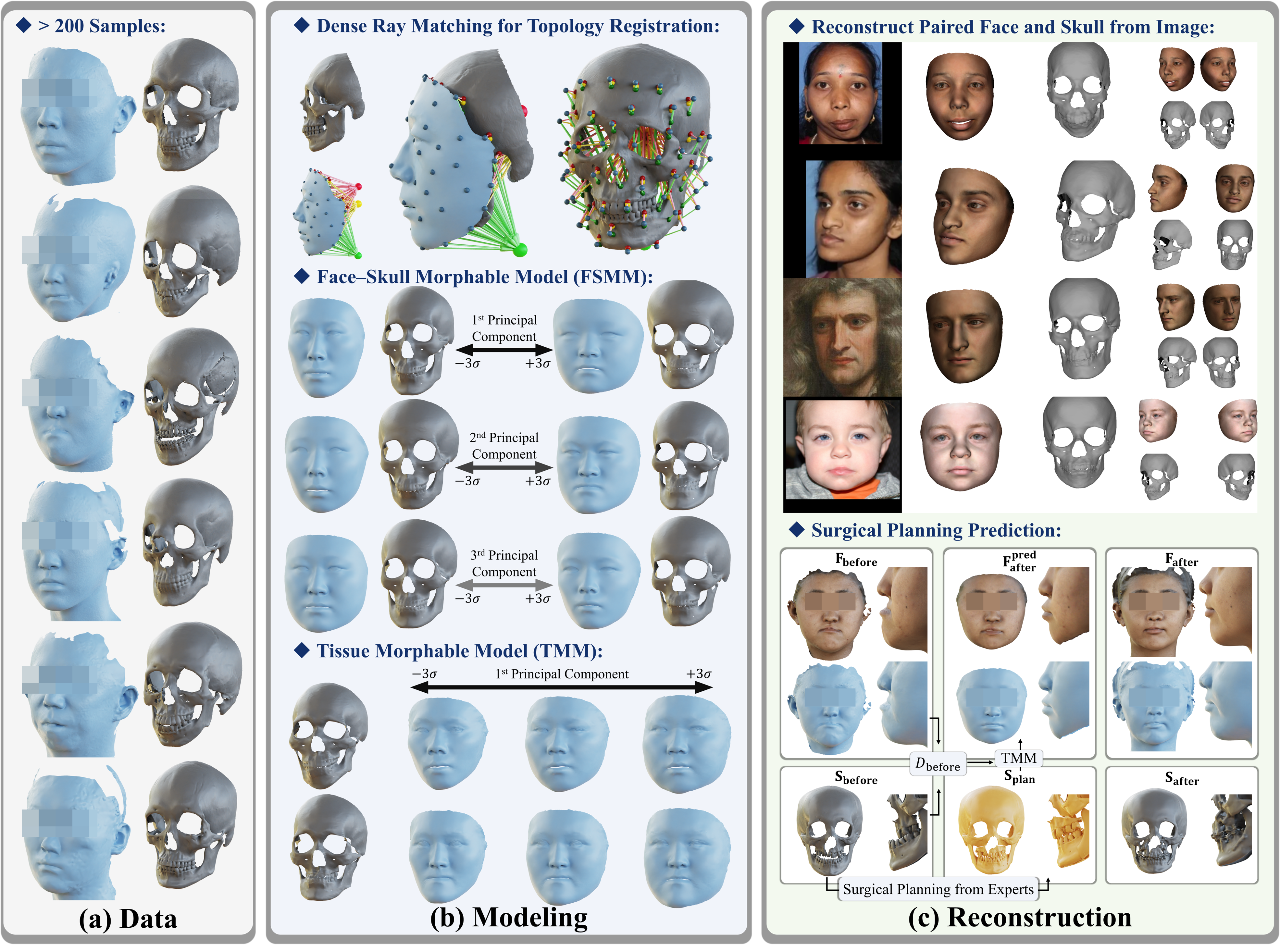}
\caption{We develop bidirectional face–skull morphable models in (a) data, (b) modeling, and (c) reconstruction.}
\label{teaser}
\end{figure}

\section{Introduction}

3D Morphable Models (3DMMs) \cite{blanz1999morphable, blanz2003face, FLAME:SiggraphAsia2017} have been widely used for facial geometry modeling, but they usually focus on the external surface, limiting their application in fields where both external appearance and internal anatomy are crucial, such as medical diagnosis, forensic analysis, surgical planning, virtual reality, and physically based facial simulation. A major obstacle in developing bidirectional face–skull morphable models lies in three key areas: data limitations, modeling difficulties, and limited application exploration.

Limited paired face–skull data is a major challenge in recent efforts to jointly model face and skull geometry, with many works relying on datasets containing fewer than 50 samples \cite{luthi2009hierarchical, paysan2009face, madsen2018probabilistic,10791299SkulltoFace,achenbach2018multilinear}. Some methods \cite{madsen2018probabilistic} even construct joint morphable models from unpaired data, which may introduce biases at the modeling stage and limit anatomical reliability. In addition, most existing models, trained on typical facial anatomy, generalize poorly to craniofacial deformities, leading to underrepresentation and limited access to inclusive modeling technologies. As a result, current 3DMM-based systems \cite{blanz1999morphable, FLAME:SiggraphAsia2017, SMPL:2015} risk perpetuating exclusion rather than promoting equity in healthcare and digital media. To overcome the data limitations, we construct a new dataset comprising over 200 high-quality samples, including normal cases, common craniofacial deformities such as cleft lip and hemifacial microsomia, and rare craniofacial conditions such as progressive hemifacial atrophy and meningoencephalocele. Every sample includes a CT-based skull, a CT-based face, and a high-fidelity textured face scan, making it the most comprehensive dataset of its kind to date, as shown in Figure~\ref{teaser}(a).

Modeling difficulties arise from both the complex surface geometry of the skull and the challenge of accurately capturing the correspondence between facial soft tissue and underlying bone. Skull deformation for topology registration is more challenging than for the face, as existing non-rigid methods \cite{nicp_github, amberg2007optimal} depend on precise control points to avoid local optima, yet identifying corresponding landmarks between source and target skull data is much harder than on the face. In many cases, skull geometry is non-watertight due to internal holes, which hinders the use of implicit representations such as signed distance fields (SDFs) \cite{mildenhall2020nerf,wang2021neus}. Although several works \cite{yang2024learning, yang2023implicit, yang2022implicit, zoss2019accurate, chandran2024anatomically} explore implicit modeling, they tend to oversimplify skull geometry, which makes them unsuitable for clinical scenarios that require high anatomical fidelity. Furthermore, the modeling of soft tissue between the face and skull remains an underexplored problem. Only a few studies \cite{achenbach2018multilinear, 10791299SkulltoFace, keller2024hit, keller2023skin, keller2022osso} have attempted to capture the tissue relationship between face and skull. These methods typically adopt a one-way mapping, either from face to skull or from skull to face, and are rarely validated on cases involving craniofacial deformities. As shown in Figure~\ref{teaser}(b), to overcome the modeling difficulties, we propose a novel dense ray matching registration method tailored for face and skull geometry. It fully incorporates the geometrical priors of anatomically aligned face and skull structures, including both normal and deformed cases, ensuring topological consistency across the face, skull, and their connecting tissues. Leveraging both the anatomical diversity of our dataset and the innovations in registration modeling, we introduce the 3D Bidirectional Face–Skull Morphable Model (BFSM). BFSM comprises two key components: the Face–Skull Morphable Model (FSMM), which enables bidirectional shape inference between the face and skull through a shared latent space, and the Tissue Morphable Model (TMM), which models tissue thickness variation to support one-to-many face reconstructions from a given skull, accounting for individual differences such as soft tissue volume and body fat distribution.

Limited exploration of joint face–skull modeling applications has constrained progress in remote healthcare, surgical planning, and other scenarios. Existing methods \cite{achenbach2018multilinear, 10791299SkulltoFace} often fail to reconstruct face–skull pairs from images alone, limiting their clinical utility, especially in cases where minimizing CT radiation exposure is critical. Additionally, surgical applications of face–skull modeling are still underexplored. As shown in Figure~\ref{teaser}(c), to address this gap, our proposed BFSM enables accurate reconstruction of face–skull pairs from a single image and predicts postoperative facial outcomes based on planned cranial modifications. Note that the prediction of surgical outcomes is based entirely on preoperative data, and results demonstrate strong consistency with actual postoperative outcomes, showcasing the potential of our model in clinical applications. These capabilities not only facilitate visualization of surgical results but also enhance doctor–patient communication, leading to more informed clinical decision-making. Our main contributions are as follows:

(1) \textbf{Data:} A dataset comprising over 200 samples, including normal and craniofacial deformity cases, with paired CT-based skulls, CT-based faces, and high-fidelity textured face scans, making it the most comprehensive dataset to date.

(2) \textbf{Modeling:} A novel dense ray matching topology registration method that uses anatomically aligned priors to ensure topological consistency across normal and deformed face–skull pairs and their tissues, leading to a novel 3D Bidirectional Face–Skull Morphable Model (BFSM) that enables bidirectional shape inference between face and skull through a shared latent space and captures tissue thickness variability to support diverse face reconstructions.

(3) \textbf{Reconstruction:} BFSM has been demonstrated in applications like remote healthcare and surgical planning, showcasing its potential to enhance doctor–patient communication and support more informed clinical decision-making. Experiments show that BFSM outperforms existing methods. BFSM will be publicly available.

\section{Related Work}

\myparagraph{Data.} Common 3D face datasets include FaceVerse \cite{wang2022faceverse}, FaceScape \cite{yang2020facescape}, NPHM \cite{giebenhain2023nphm}, and LYHM \cite{dai2020statisticalLYHM}. While some studies offer 3D skull data \cite{madsen2018probabilistic,achenbach2018multilinear,luthi2009hierarchical, paysan2009face}, the number of samples is significantly smaller than that of face datasets, and paired face–skull data remains particularly scarce. Existing paired datasets are typically limited in scale, lack sample diversity, and do not include pathological cases.

\myparagraph{Registration and Modeling.} Deformation for topology registration is essential for modeling 3D face–skull geometry. Existing non-rigid deformation methods \cite{nicp_github, amberg2007optimal, porto2021alpaca}, mostly based on nearest-neighbor strategies, rely heavily on accurate control points to avoid local optima. In practice, only a small number of sparse landmarks are available, and even slight errors can cause significant registration failures. 

Widely used morphable models include the Basel Face Model (BFM) \cite{blanz1999morphable, blanz2003face} and FLAME \cite{FLAME:SiggraphAsia2017}. Several works explore implicit representations \cite{yang2024learning, yang2023implicit, yang2022implicit, zoss2019accurate, chandran2024anatomically} for physically based facial simulation, but often simplify skull geometry to a degree that limits clinical relevance. Other methods \cite{keller2024hit, keller2023skin, keller2022osso, achenbach2018multilinear, 10791299SkulltoFace} attempt to model face–skull relationships jointly, yet they typically overlook pathological cases, which may reinforce exclusion and hinder equitable healthcare and representation.

\begin{figure}[t]
\centering
\includegraphics[width=0.98\columnwidth]{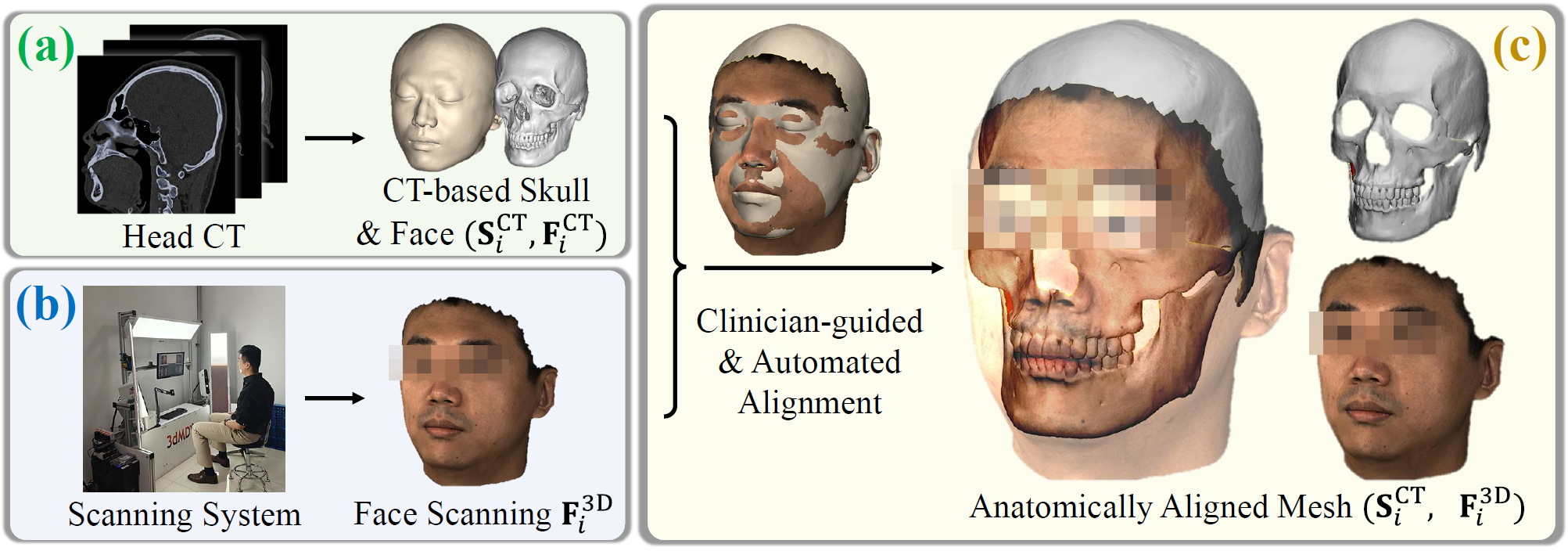}
\caption{(a) and (b): Overview of our face and skull data acquisition process. (c): Clinician-guided and automated alignment of CT and scanning data.}
\label{preprocess}
\end{figure}

\begin{figure*}[t]
\centering
\includegraphics[width=0.98\textwidth]{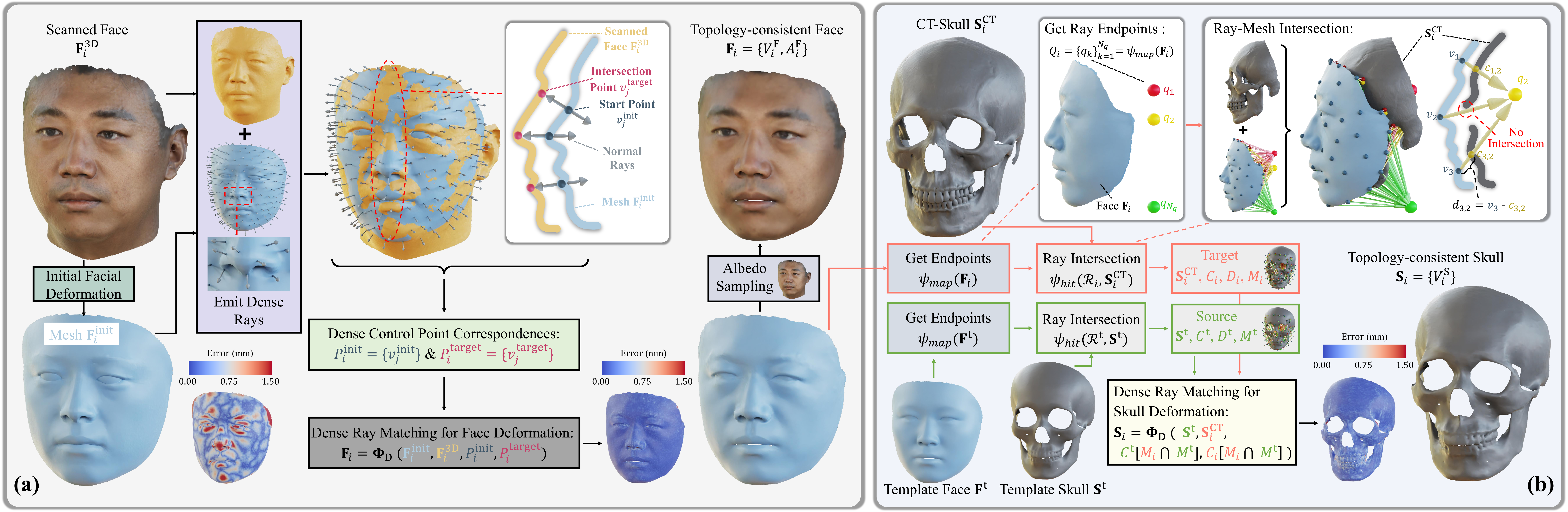} 
\caption{Overview of our topology-consistent deformation pipeline. (a) Dense ray matching for face mesh deformation. (b) Dense ray matching for skull mesh deformation.}
\label{dense-matching}
\end{figure*}

\myparagraph{Reconstruction and Applications.}
Many methods \cite{wang2025srmhair, wang20243d, wang2024s2td, DECA:Siggraph2021, SMIRK, deng2019accurate} achieve single-image face reconstruction. However, reconstructing both face and skull from a single image is rarely explored, limiting use in remote healthcare and CT-limited scenarios. Existing digital human editing techniques \cite{potamias2024shapefusion, tarasiou2024locally, foti20223d} mainly target facial surfaces and neglect clinical needs for joint face–skull manipulation.

\section{Modeling}
In this section, we first describe our data acquisition process. Next, we propose a novel dense ray-matching deformation method to achieve topology-consistent face and skull registration and establish point-to-point anatomical tissue correspondences. Finally, we present the Face–Skull Morphable Model and the Tissue Morphable Model, which together form the Bidirectional Face–Skull Morphable Model (BFSM). Applications of BFSM in remote healthcare and surgical planning are detailed in detail in Section~\ref{4Reconstruction}.

\subsection{Data Acquisition and Preprocessing}

\myparagraph{Data Acquisition.} We collaborate closely with several experienced clinicians, with over a decade of clinical practice. These experts meticulously review the 3D data for each case to ensure anatomical accuracy and clinical relevance. Under their guidance, we filter out anatomically irrelevant or redundant structures—such as internal cranial tissues or scattered noise artifacts. As shown in Figure~\ref{preprocess} (a) and (b), our dataset is composed of two sources: computed tomography (CT) and 3D textured scanning. The CT data, encompassing both skull $\mathbf{S}_{i}^\mathrm{CT}$ and face $\mathbf{F}_{i}^{\mathrm{CT}}$, are acquired using a \textit{Philips Brilliance iCT 128-slice spiral CT scanner}, with scanning focused on the craniofacial regions. High-resolution textured face scannings $\mathbf{F}_{i}^\mathrm{3D}$ are captured using the \textit{3dMD Face System}. The facial scans cover the area from ear to ear horizontally and from the hairline down to the neck vertically, with texture at a resolution of 5 megapixels. The meshes $\mathbf{S}_{i}^{\mathrm{CT}}$, $\mathbf{F}_{i}^{\mathrm{CT}}$, and $\mathbf{F}_{i}^{\mathrm{3D}}$ represent the CT-based skull, CT-based face, and 3D-scanned face of subject $i$, respectively.

\myparagraph{Clinician-guided and Automated Alignment.} Since $\mathbf{F}_{i}^{\mathrm{CT}}$ lacks texture information and is typically captured with the subject's eyes closed, it is not ideal for directly constructing a morphable facial model. Therefore, we use $\mathbf{F}_{i}^{\mathrm{CT}}$ as an intermediate modality to anatomically align $\mathbf{F}_{i}^{\mathrm{3D}}$ with $\mathbf{S}_{i}^{\mathrm{CT}}$. Specifically, clinical experts first manually annotate anatomical landmarks on both $\mathbf{F}_{i}^{\mathrm{CT}}$ and $\mathbf{F}_{i}^{\mathrm{3D}}$ to perform an initial rigid alignment. This is followed by automated alignment using professional-grade medical modeling software, such as \textit{Materialise 3-matic Medical}. The registration results are then reviewed and refined by experienced clinicians to ensure anatomically accurate alignment between $\mathbf{F}_{i}^{\mathrm{3D}}$ and $\mathbf{S}_{i}^{\mathrm{CT}}$. An overview of this process is shown in Figure~\ref{preprocess}(c).

\myparagraph{Distribution.} Our dataset contains over 200 samples and consists of two major categories: normative (\textit{i.e.}, anatomically normal) and craniofacial deformity cases. The deformity group includes patients diagnosed with conditions such as cleft lip, Treacher Collins syndrome, hemifacial microsomia, meningoencephalocele, and progressive hemifacial atrophy. Some examples are shown in Figure~\ref{teaser}(a), where the facial textures of $\mathbf{F}_{i}^{\mathrm{3D}}$ have been removed to protect subject privacy. Statistics regarding the data distribution are summarized in supplemental materials.

\myparagraph{Ethics.} Ethical approval was obtained from the Institutional Review Board, and all participants provided written informed consent. The study adhered to the principles of the \textit{Declaration of Helsinki}.

\subsection{Dense Ray Matching for Topology Registration}

Our goal is to deform a pair of fixed-topology face and skull templates (\( \mathbf{F}^{\mathrm{t}} \) and \( \mathbf{S}^{\mathrm{t}} \)) to match the anatomically aligned target data (\( \mathbf{F}_i^{\mathrm{3D}} \) and \( \mathbf{S}_i^{\mathrm{CT}} \)). This yields topology-consistent face \( \mathbf{F}_i \) and skull \( \mathbf{S}_i \), along with dense tissue correspondences $D_i$ between them.

We first perform non-rigid registration on the facial surface and then leverage its strong geometric priors to guide the topology skull registration, resulting in dense tissue correspondences between the face and the skull. To support this, we introduce a novel dense ray-matching deformation method that establishes dense and anatomically consistent vertex correspondences for the non-rigid deformation module. Existing non-rigid deformation methods \cite{nicp_github, amberg2007optimal}, which are predominantly based on nearest-neighbor strategies, often rely heavily on accurately estimated control points to avoid convergence to local optima. Our approach demonstrates that our dense ray-matching-based guidance significantly improves registration robustness, even in challenging cases involving craniofacial deformities. Comprehensive quantitative evaluations are provided in the experiment section.

\myparagraph{Initial Facial Deformation.}
For initialization, we first register the target face \( \mathbf{F}_i^{\mathrm{3D}} \) using BFM \cite{blanz1999morphable} with differentiable rendering \cite{Laine2020diffrast, ravi2020pytorch3d}. Specifically, the 3D facial scan \( \mathbf{F}_i^{\mathrm{3D}} \) is rendered to obtain its corresponding RGB image, depth map, and silhouette. We extract predicted facial attributes such as landmarks, depth, texture, and semantic part segmentation \cite{wang20243d} for supervision signals to iteratively optimize the coefficients of the BFM. The result of this optimization is a topology-consistent face mesh \( \mathbf{F}^{\mathrm{init}}_{i} \). Figure~\ref{dense-matching}(a) shows that \( \mathbf{F}^{\mathrm{init}}_{i} \) is coarsely aligned with \( \mathbf{F}_i^{\mathrm{3D}} \), with the alignment error visualized as a heatmap.

\myparagraph{Deformation Module \( \boldsymbol{\Phi}_\mathrm{D} \).} We define a non-rigid deformation module \( \boldsymbol{\Phi}_\mathrm{D} \) that deforms a source mesh \( \mathbf{X} \) to a target mesh \( \mathbf{X}' \), guided by control points. The module takes two 3D meshes \( \mathbf{X} \) and \( \mathbf{X}' \), and their corresponding sets of control points \( {P} \in \mathbb{R}^{k \cdot 3} \) and \( {P}'\in \mathbb{R}^{k \cdot 3} \) as input. Note that $\mathbf{X}$ and $\mathbf{X}'$ differ in their topological structures. \( \boldsymbol{\Phi}_\mathrm{D} \) outputs a registered mesh \( \mathbf{X}_{\mathrm{r}} \):
\begin{equation}
\begin{aligned}
\begin{array}{l}
\mathbf{X}_{\mathrm{r}} = \boldsymbol{\Phi}_\mathrm{D}(\mathbf{X}, \mathbf{X}' , {P}, {P}')
\end{array}
\end{aligned},
\label{Basic_Deformation_Module}
\end{equation}
where the deformation module \( \boldsymbol{\Phi}_\mathrm{D} \) is defined as the solution to the following optimization objective:
\begin{equation}
\small
\begin{aligned}
\mathcal{L}_{\text{D}} &= \mathcal{L}_{\text{mesh}} + \lambda \mathcal{L}_{\text{smooth}} + \mu \mathcal{L}_{\text{P}}\\
 &=\sum_{i=1}^{N} \left\| \mathbf{T}({v}_i) - \Pi({v}_i, \mathbf{X}') \right\|^2  \\
 & \quad+ \lambda \sum_{i=1}^{N} \left\| \Delta \mathbf{T}(\mathbf{v}_i) \right\|^2  + \mu \left\| \mathbf{T}({P}) - {P}' \right\|^2,
\end{aligned}
\label{Basic_Deformation_Module_Loss}
\end{equation}
where the deformation \( \mathbf{T}(\cdot) \) is optimized as a vertex affine field using the AMSGrad \cite{reddi2019convergence} optimizer within a PyTorch \cite{paszke2019pytorch} framework \cite{nicp_github, amberg2007optimal}. $\{v_i\}_{i=1}^{N}$ are the vetices in mesh \(\mathbf{X}\). $\mathcal{L}_{\text{mesh}}$ encourages mesh alignment between the source and the target via nearest-neighbor matching, \( \Pi({v}_i, \mathbf{X}') \) finds the closest point on the target mesh \( \mathbf{X}' \) to the vertex \( {v}_i \). The smoothness term \( \mathcal{L}_{\text{smooth}} \) enforces Laplacian regularity \cite{amberg2007optimal} to ensure locally smooth deformations, where \( \Delta \) is the mesh Laplacian operator. The control point term $\mathcal{L}_{\text{P}}$ ensures anatomical consistency at landmark locations.

\( \boldsymbol{\Phi}_\mathrm{D} \) relies on accurate control point correspondences to avoid local optima and achieve topology-consistent registration, practical scenarios often provide only a few dozen sparse landmarks on the face or skull. Due to the sparsity of control points, even minor inaccuracies can lead to significant registration errors. We therefore propose a dense ray-matching-based guidance to extract tens of thousands of anatomically plausible correspondences, significantly improving the effectiveness and robustnes of \( \boldsymbol{\Phi}_\mathrm{D} \).

\myparagraph{Dense Ray Matching for Face Deformation.} As shown in Figure~\ref{dense-matching}(a), starting from the surface of \( \mathbf{F}_i^{\mathrm{init}} \), we emit two rays from each vertex \( {v}_j^{\mathrm{init}} \in \mathbf{F}_i^{\mathrm{init}} \), following the directions of the positive and negative vertex normals. These rays, visualized in Figure~\ref{dense-matching}(a), intersect the target scan \( \mathbf{F}_i^{\mathrm{3D}} \) with each intersection point denoted as \( {v}_j^{\mathrm{target}} \). Ray origins and intersections construct two dense point sets, ${P}_i^{\mathrm{init}} = \{ {v}_j^{\mathrm{init}} \}$ and ${P}_i^{\mathrm{target}}  = \{ {v}_j^{\mathrm{target}} \}$, establishing dense control point correspondences between \( \mathbf{F}_i^{\mathrm{init}} \) and \( \mathbf{F}_i^{\mathrm{3D}} \). These correspondences are used as supervision signals for the module \( \boldsymbol{\Phi}_\mathrm{D} \):
\begin{equation}
\begin{aligned}
\begin{array}{l}
 \mathbf{F}_i  = \boldsymbol{\Phi}_\mathrm{D}(\mathbf{F}_i^{\mathrm{init}}, \mathbf{F}_i^{\mathrm{3D}} , {P}_i^{\mathrm{init}}, {P}_i^{\mathrm{target}})
\end{array}
\end{aligned},
\label{face_Deformation}
\end{equation}
where we obtain the topology-consistent face mesh \( \mathbf{F}_i = \{ {V}_i^\mathrm{F}, {A}_i^\mathrm{F} \} \) with fixed triangle-faces. \( {V}_i^\mathrm{F} \in \mathbb{R}^{N_\mathrm{F} \cdot 3} \) represents the vertices and \( {A}_i^\mathrm{F} \) denotes the per-vertex albedo. The alignment accuracy between \( \mathbf{F}_i \) and the input scan \( \mathbf{F}_i^{\mathrm{3D}} \) is visualized as an error heatmap in Figure~\ref{dense-matching}(a), showing that \( \mathbf{F}_i \) closely matches the geometry of \( \mathbf{F}_i^{\mathrm{3D}} \). In rare cases where the normal rays fail to intersect the target facial mesh $ \mathbf{F}_i^{\mathrm{3D}} $, the nearest point to \( {v}_j^{\mathrm{init}} \) on \( \mathbf{F}_i^{\mathrm{3D}} \) is used as \( {v}_j^{\mathrm{target}} \).

\myparagraph{Dense Ray Matching for Skull Deformation.} As shown in Figure~\ref{dense-matching}(b), based on the face deformation step, we have a pair of anatomically aligned meshes \( \mathbf{F}_i \) and \( \mathbf{S}_i^{\mathrm{CT}} \). Our goal is to deform and register the skull template \( \mathbf{S}^{\mathrm{t}} \) from a pair of fixed-topology face and skull templates (\( \mathbf{F}^{\mathrm{t}} \) and \( \mathbf{S}^{\mathrm{t}} \)) to align with \( \mathbf{S}_i^{\mathrm{CT}} \). Note that \( \mathbf{F}_i \) and \( \mathbf{F}^{\mathrm{t}} \) share the same topology, whereas \( \mathbf{S}_i^{\mathrm{CT}} \) and \( \mathbf{S}^{\mathrm{t}} \) do not. Due to the more complex geometry of the skull, including holes and other anatomical structures, we adopt a dense ray matching strategy to leverage the structural relationship between the face and the skull. This allows us to obtain the control points needed to deform the skull template \( \mathbf{S}^{\mathrm{t}} \) to the target CT skull \( \mathbf{S}^{\mathrm{CT}} \) in the deformation module \( \boldsymbol{\Phi}_\mathrm{D} \) and establish dense tissue correspondences between the face and the skull.

Given two anatomically aligned pairs (\( \mathbf{F}_i \), \( \mathbf{S}_i^{\mathrm{CT}} \)) and (\( \mathbf{F}^{\mathrm{t}} \), \( \mathbf{S}^{\mathrm{t}} \)), we first construct a fixed mapping function \( \psi_\mathrm{map}(\cdot) \) that maps facial landmarks on the face mesh to a series of spatially meaningful 3D points. These points are located in or near the occipital region of the head and serve as the endpoints of dense rays used to deform the skull. We apply \( \psi_\mathrm{map}(\cdot) \) to \( \mathbf{F}_i \) and \( \mathbf{F}^{\mathrm{t}} \), respectively, obtaining two sets of endpoints \( Q_i = \{ {q}_k \}_{k=1}^{N_q} \) and \( Q^{\mathrm{t}} = \{ {q}_k^{\mathrm{t}} \}_{k=1}^{N_q} \). For the subject $i$ and template face meshes \( \mathbf{F}_i \) and \( \mathbf{F}^{\mathrm{t}} \), their vertex sets are denoted as \( {V}_i^{\mathrm{F}} = \{ {v}_j \}_{j=1}^{N_\mathrm{F}} \in \mathbf{F}_i \) and \( {V}^{\mathrm{t}} = \{ {v}_j^{\mathrm{t}} \}_{j=1}^{N_\mathrm{F}} \in \mathbf{F}^{\mathrm{t}} \), respectively. For each mesh, we construct a set of rays \( \mathcal{R}_i = \{ \mathbf{r}_{j,k} \} \) and \( \mathcal{R}^{\mathrm{t}} = \{ \mathbf{r}_{j,k}^{\mathrm{t}} \} \), where each ray originates from a facial vertex \( {v}_j \) or \( {v}_j^{\mathrm{t}} \) and points toward a endpoint \( {q}_k \in Q_i \) or \( {q}_k^{\mathrm{t}} \in Q^{\mathrm{t}} \), as shown in  Figure~\ref{dense-matching}(b).

Since each ray in \( \mathcal{R}_i \) and \( \mathcal{R}^{\mathrm{t}} \) has an ordered pair of origin and end points, the corresponding intersection points with the skull meshes \( \mathbf{S}_i^{\mathrm{CT}} \) and \( \mathbf{S}^{\mathrm{t}} \) are also ordered accordingly. We introduce a ray-mesh intersection function, denoted as \( \psi_{\mathrm{hit}}(\cdot) \), to compute these intersections on the skulls:
\begin{equation}
\begin{aligned}
C_i, D_i, M_i &= \psi_{\mathrm{hit}}(\mathcal{R}_i, \mathbf{S}_i^{\mathrm{CT}})\\C^{\mathrm{t}}, D^{\mathrm{t}}, M^{\mathrm{t}} &= \psi_{\mathrm{hit}}(\mathcal{R}^{\mathrm{t}}, \mathbf{S}^{\mathrm{t}})
\end{aligned}
\label{dense_skull_ray},
\end{equation}
where \( C_i = \{c_{j,k}\} \in \mathbb{R}^{(N_\mathrm{F} \cdot N_q) \cdot 3} \) and \( C^{\mathrm{t}} = \{{c_{j,k}}^{\mathrm{t}}\} \in \mathbb{R}^{(N_\mathrm{F} \cdot N_q) \cdot 3} \)  represent the intersection points on skull. \( D_i = \{d_{j,k}\} \in \mathbb{R}^{(N_\mathrm{F} \cdot N_q) \cdot 3} \) and \( D^{\mathrm{t}} \in \mathbb{R}^{(N_\mathrm{F} \cdot N_q) \cdot 3} \) denote the direction vectors from the skull to the face with distance and \(d_{j,k} = v_j - c_{j,k}\). \( D_i \) is a dense estimation of the tissue thickness between the face and the skull, which is later used for constructing the Tissue Morphable Model. The masks \( M_i, M^{\mathrm{t}} \in \{0,1\}^{N_\mathrm{F} \cdot N_q} \) indicate the validity of each intersection. Due to anatomical features such as holes and discontinuities in the skull geometry, some rays \( \mathbf{r}_{j,k} \in \mathcal{R}_i \) or \( \mathbf{r}_{j,k}^{\mathrm{t}} \in \mathcal{R}^{\mathrm{t}} \) may not produce valid intersections. In such cases, \( \psi_{\mathrm{hit}}(\cdot) \) assigns a sentinel value and marks these entries as invalid in the corresponding masks \( M_i \) and \( M^{\mathrm{t}} \).

At this stage, we have acquired the dense skull control points required by the deformation module \( \boldsymbol{\Phi}_\mathrm{D} \) for the pairs \( (\mathbf{F}_i, \mathbf{S}_i^{\mathrm{CT}}) \) and \( (\mathbf{F}^{\mathrm{t}}, \mathbf{S}^{\mathrm{t}}) \), enabling the deformation of the skull template \( \mathbf{S}^{\mathrm{t}} \) to the target skull \( \mathbf{S}_i^{\mathrm{CT}} \):
\begin{equation}
\begin{aligned}
\begin{array}{l}
 \mathbf{S}_i  = \boldsymbol{\Phi}_\mathrm{D}(\mathbf{S}^{\mathrm{t}}, \mathbf{S}_i^{\mathrm{CT}},  C^{\mathrm{t}} [M_i \cap M^{\mathrm{t}}], C_i[M_i \cap M^{\mathrm{t}}])
\end{array}
\end{aligned},
\label{skull_Deformation}
\end{equation}
where we obtain the topology-consistent skull mesh \( \mathbf{S}_i = \{ {V}_i^\mathrm{S} \} \) with fixed triangle-faces. \( {V}_i^\mathrm{S} \in \mathbb{R}^{N_\mathrm{S} \cdot 3} \). As shown in Figure~\ref{dense-matching}(b), the template pair is fixed. For each subject \(i\), we apply the non-rigid deformation process based on dense ray matching as defined in Equations~\ref{face_Deformation} and~\ref{skull_Deformation}. This results in a collection of \(N_{\text{sample}}\) topologically aligned face–skull mesh pairs \(\{ \mathbf{F}_i, \mathbf{S}_i \}_{i=1}^{N_{\text{sample}}}\), along with corresponding dense tissue thickness \(\{ D_i \}_{i=1}^{N_{\text{sample}}}\) between face and skull. The impact of ray origin and endpoint numbers (\(N_O\) and \(N_q\)) on registration accuracy is analyzed in the Ablation Study \ref{AblationStudy}.

To support Tissue Morphable Model in the next section,  
we construct a fixed index mapping \( \mathcal{I}_{\mathrm{t}} \in \mathbb{R}^{N_\mathrm{F} \cdot N_q} \) between the template skull vertices \( V^{\mathrm{t}}_\mathrm{S} \in \mathbb{R}^{N_\mathrm{S} \cdot 3} \) and the control points \( C^{\mathrm{t}} \in \mathbb{R}^{(N_\mathrm{F} \cdot N_q) \times 3} \),  
where each index is determined by nearest-neighbor search from \( C^{\mathrm{t}} \) to \( V^{\mathrm{t}}_\mathrm{S} \). \( \mathcal{I}_{\mathrm{t}} \) is later used to locate the origin vertices for the tissue thickness vectors on any given skull mesh.

\begin{figure}[t]
\centering
\includegraphics[width=0.98\columnwidth]{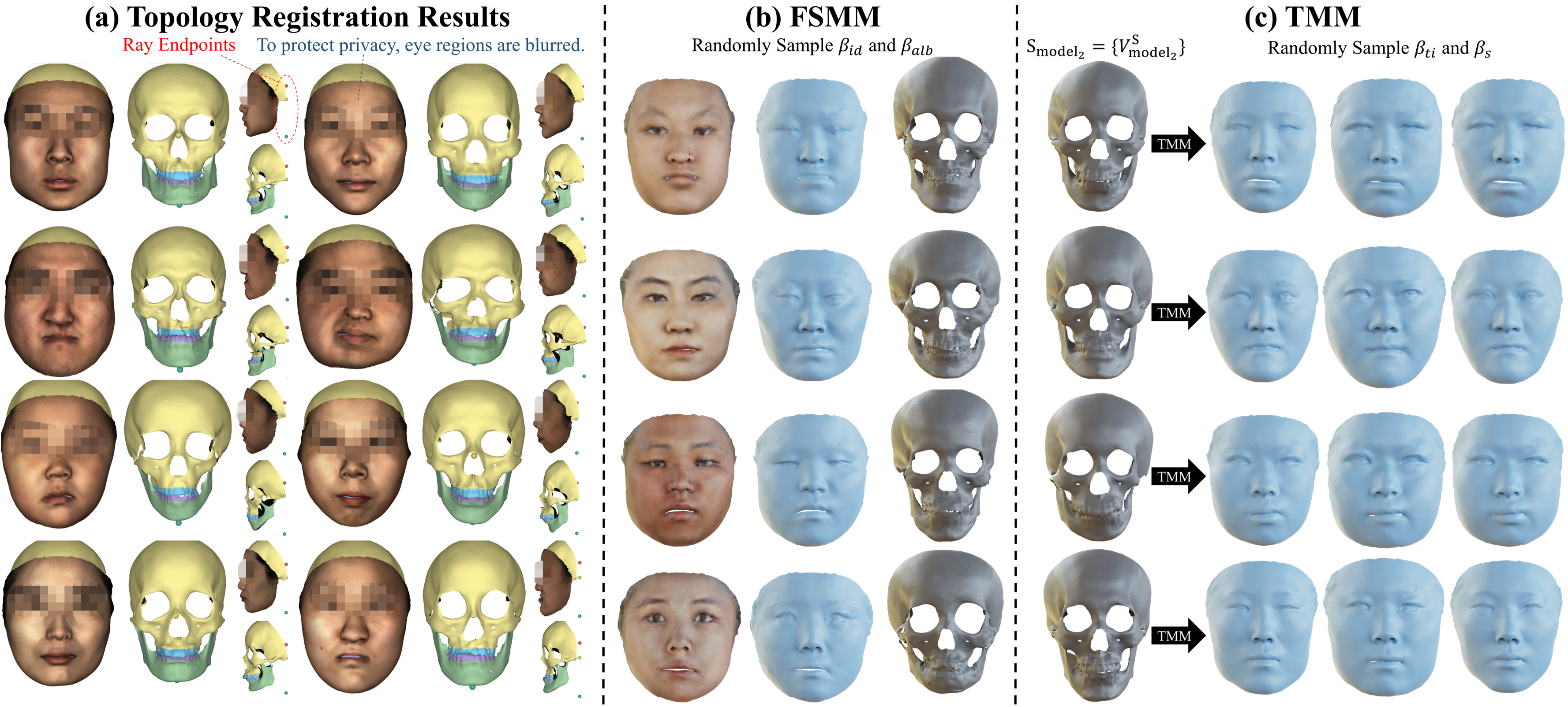}
\caption{Results visualization of dense ray matching for topology registration (a), FSMM (b), and TMM (c).}
\label{Visualization}
\end{figure}

\subsection{Bidirectional Face-Skull Morphable Model}

To enhance robustness, each subject \( i\) is horizontally flipped, yielding \( N_{\text{sample}} = 161 \cdot 2 = 322 \) samples.

\myparagraph{Face–Skull Morphable Model.} 
Based on prior work on morphable models~\cite{wood2021fake, blanz1999morphable,wang2025srmhair}, and Principal Component Analysis (PCA)~\cite{Sirovich:87, jolliffe2016principal}, we construct a {3D Face--Skull Morphable Model (FSMM)}. FSMM enables the generation of a new pair of anatomically aligned face and skull from a {shared set of coefficients}:
\begin{equation}
\begin{aligned}
V_\mathrm{model_1}^{\mathrm{FS}} &= \left[ V_\mathrm{model_1}^\mathrm{F} \,\middle\|\, V_\mathrm{model_1}^\mathrm{S} \right] \\ &= \boldsymbol{R}({{\beta} _{a}})({\overline V}  + {{\beta} _{id}}{\boldsymbol{B}_{id}}) + {{\beta} _{t}}\\
A_\mathrm{model_1}^\mathrm{F} &= {\overline A}  + {{\beta} _{alb}}{\boldsymbol{B}_{alb}}
\end{aligned}\quad,
\label{FSMM-equ}
\end{equation}
where \( {\beta}_{id} \in \mathbb{R}^{n_{id}} \), \( {\beta}_{alb} \in \mathbb{R}^{n_{alb}} \), \( {\beta}_{a} \in \mathbb{R}^{3} \), and \( {\beta}_{t} \in \mathbb{R}^{3} \) are the identity, albedo, rotation, and translation coefficients, respectively. \( \boldsymbol{R}({\beta}_{a}) \in \mathbb{R}^{3 \cdot 3} \) is the rotation matrix corresponding to angles \( {\beta}_{a} \in \mathbb{R}^{3} \). \( \overline{V} \in \mathbb{R}^{(N_{\mathrm{F}} + N_{\mathrm{S}}) \cdot 3} \) and \( \overline{A} \in \mathbb{R}^{N_{\mathrm{F}} \cdot 3} \) denote the mean shape and mean albedo, computed from the face-skull joint vertex set \( \left[ V_{i}^\mathrm{F} \,\middle\|\, V_{i}^\mathrm{S} \right]_{i=1}^{N_{\text{sample}}} \) and the facial texture set \( \{ A_i^{\mathrm{F}} \}_{i=1}^{N_{\text{sample}}} \), respectively. \( \boldsymbol{B}_{id} \) and \( \boldsymbol{B}_{alb} \) are obtained by applying PCA to the concatenated vertex pairs \( \left[ V_{i}^\mathrm{F} \,\middle\|\, V_{i}^\mathrm{S} \right]_{i=1}^{N_{\text{sample}}} \) and the facial albedo set \( \{ A_i^{\mathrm{F}} \}_{i=1}^{N_{\text{sample}}} \), respectively. The operator \( \left[ \cdot \,\middle\|\, \cdot \right] \) denotes vertex concatenation.

\myparagraph{Tissue Morphable Model.} In real-world scenarios, a single skull may correspond to multiple facial appearances due to factors like age or body fat. To capture this one-to-many relationship, we construct a Tissue Morphable Model (TMM) based on tissue thickness vectors \( \{ D_i \}_{i=1}^{N_{\text{sample}}} \). The tissue depth from skull to face can be generated by:
\begin{equation}
{D}_{\mathrm{model_2}} = \left( \overline{{D}} + {{\beta} _{ti}}{\boldsymbol{B}_{ti}} \right) \cdot \beta_s
\label{TMM}\quad,
\end{equation}
where \( \overline{D} \in \mathbb{R}^{(N_{\mathrm{F}} \cdot N_q) \times 3} \) denotes the mean tissue thickness vector,  \( \boldsymbol{B}_{ti}  \) represents the PCA basis of tissue thickness \( \{ D_i \}_{i=1}^{N_{\text{sample}}} \),  
\( \beta_{ti} \in \mathbb{R}^{n_{ti}} \) is the individual-specific tissue coefficients, and \( \beta_s \in \mathbb{R} \) is a global scalar controlling overall tissue thickness scaling. Given a skull mesh \( \mathbf{S}_\mathrm{model_2} = \{ V_\mathrm{model_2}^\mathrm{S} \} \) with the same fixed topology as \( \mathbf{S}^{\mathrm{t}} \), a new set of facial surface points can be generated as:
\begin{equation}
\begin{aligned}
V'_{\mathrm{model_2}} &= V_\mathrm{model_2}^\mathrm{S}\left[\mathcal{I}_{\mathrm{t}} \right] + {D}_{\mathrm{model_2}} \quad,\\
 V_\mathrm{model_2}^\mathrm{F} &= \boldsymbol{\Phi}_\mathrm{fit}(V'_{\mathrm{model_2}})
\label{fit_TMM}\quad,
\end{aligned}
\end{equation}
where \( \mathcal{I}_{\mathrm{t}} \in \mathbb{R}^{(N_{\mathrm{F}} \cdot N_q)} \) is the precomputed index map that associates each tissue thickness vector to a vertex on the given skull \( \mathbf{S}_\mathrm{model_2} \), resulting \(V_\mathrm{model_2}^\mathrm{S}\left[\mathcal{I}_{\mathrm{t}} \right]  \in \mathbb{R}^{(N_{\mathrm{F}} \cdot N_q) \times 3} \) and \(V'_{\mathrm{model_2}} \in \mathbb{R}^{(N_{\mathrm{F}} \cdot N_q) \times 3}\), and \( \boldsymbol{\Phi}_{\mathrm{fit}}(\cdot) \) denotes a face mesh reconstruction function that fits the facial surface points \( V'_{\mathrm{model_2}} \) onto a topology-consistent facial mesh. \( \boldsymbol{\Phi}_{\mathrm{fit}}(\cdot) \) can be implemented through a deformation process similar to Equations~\ref{face_Deformation} and~\ref{skull_Deformation} or using FSMM. If the given skull mesh does not share the same topology as \( \mathbf{S}^{\mathrm{t}} \), a similar deformation strategy can be applied to obtain a topology-consistent representation, which we omit here for brevity. 
Note that the facial pose in TMM is inherently determined by the input skull \( \mathbf{S}_\mathrm{model_2} \) and Equation~\ref{fit_TMM} does not require additional pose coefficients.

\myparagraph{Visualization.} Figure~\ref{Visualization}(a) shows the results of topology registration. To protect privacy, eye regions are blurred. The consistent topology of the registered skull enables straightforward segmentation into anatomical subregions. Ray endpoints based on the face mesh are derived via \( \psi_\mathrm{map}(\cdot) \). Figure~\ref{Visualization}(b) and Figure~\ref{Visualization}(c) illustrate anatomically aligned face–skull pairs generated by randomly sampling \((\beta_{{id}}, \beta_{{alb}})\) from FSMM and \((\beta_{{ti}}, \beta_s)\) from TMM, respectively.

\begin{figure}[t]
\centering
\includegraphics[width=0.98\columnwidth]{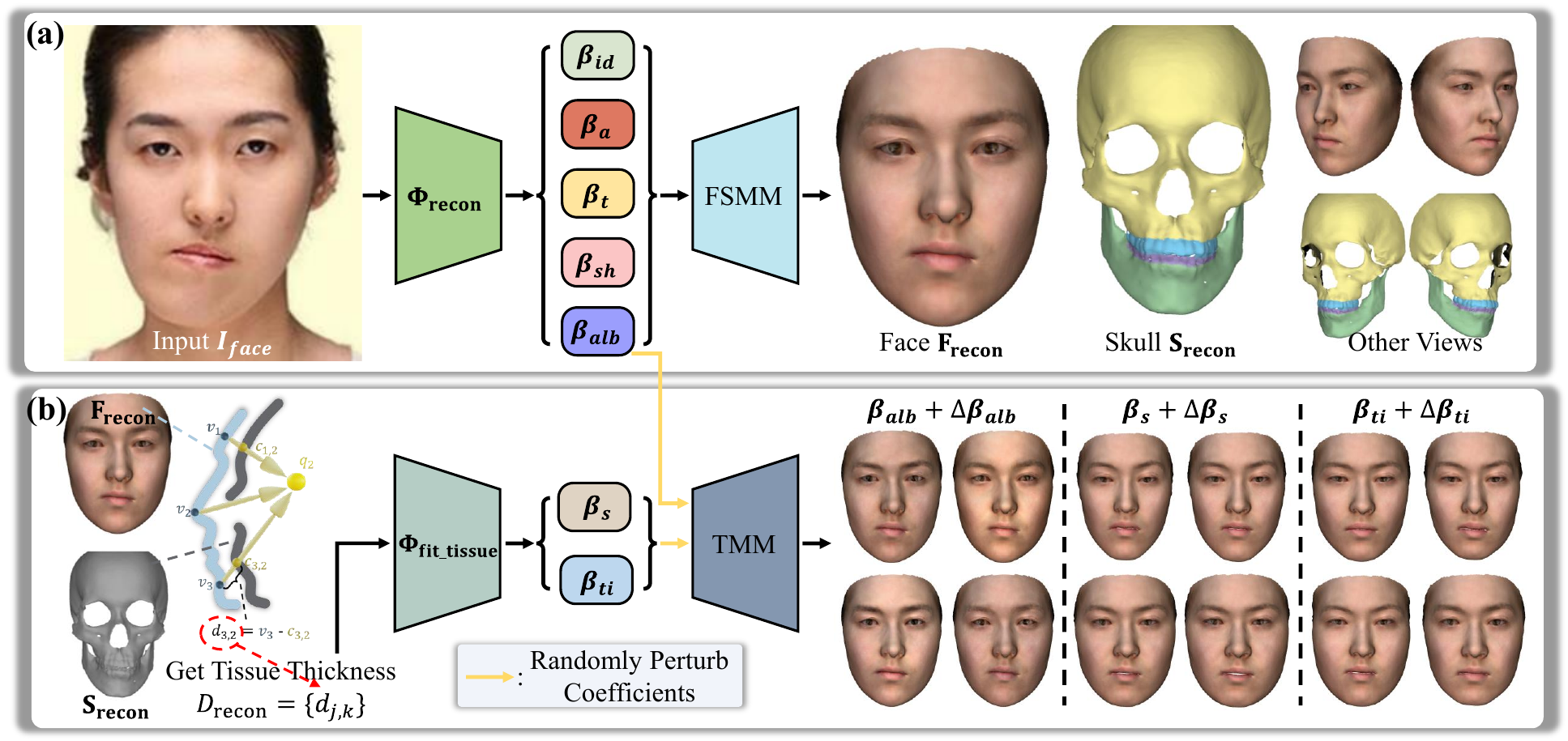}
\caption{(a) Reconstruction via FSMM. (b) Facial variation by perturbing TMM coefficients for a given skull.}
\label{Reconstruct_Paired_Face_and_Skull_from_Image}
\end{figure}

\section{Reconstruction}
\label{4Reconstruction}
\subsection{Reconstruct Paired Face and Skull from a Single Image}
\begin{figure}[t]
\centering
\includegraphics[width=0.98\columnwidth]{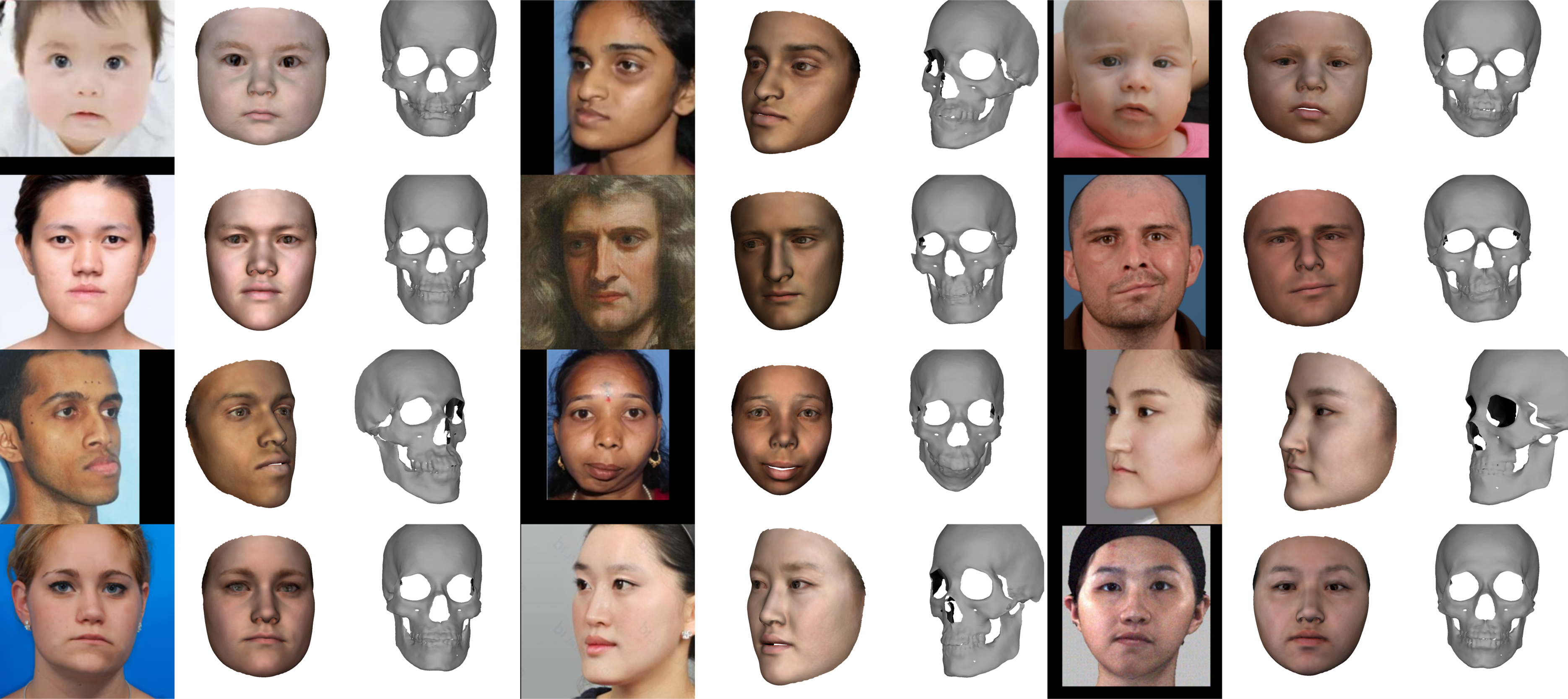}
\caption{Face–skull reconstruction from a single image.}
\label{recon_res2}
\end{figure}

Since the FSMM enables the generation of anatomically aligned face–skull pairs from a shared coefficients (\({\beta}_{id}, {\beta}_{a},{\beta}_{t},{\beta}_{alb}\)), we can leverage existing monocular 3D face reconstruction methods~\cite{wang2024s2td, wang20243d, DECA:Siggraph2021, deng2019accurate} to supervise the face component \( V_\mathrm{model_1}^\mathrm{F} \) in Equation~\ref{FSMM-equ} using only 2D facial image training data. Specifically, we train a backbone \( \boldsymbol{\Phi}_\mathrm{recon} \) to predict the coefficients of the FSMM, thereby enabling the reconstruction of an anatomically aligned face–skull pair ($\mathbf{F}_\mathrm{recon}$, $\mathbf{S}_\mathrm{recon}$) from a single input image $I_{face}$, as illustrated in Figure~\ref{Reconstruct_Paired_Face_and_Skull_from_Image}(a). Note that in Figure~\ref{Reconstruct_Paired_Face_and_Skull_from_Image}(a), \( {\beta}_{sh} \) denotes the Spherical Harmonics (SH) \cite{ramamoorthi2001efficient} coefficients used for facial appearance modeling. As shown in Figure~\ref{Reconstruct_Paired_Face_and_Skull_from_Image}(b), given the reconstructed face--skull pair \( \mathbf{F}_\mathrm{recon} \) and \( \mathbf{S}_\mathrm{recon} \), the tissue thickness vector \( D_{\mathrm{recon}} \) can be estimated using Equation~\ref{dense_skull_ray}. Based on this, we design a gradient-based optimization framework \( \boldsymbol{\Phi}_\mathrm{fit\_tissue} \) to fit the TMM coefficients \((\beta_{ti}, \beta_{s})\) corresponding to \( D_{\mathrm{recon}} \):
\begin{equation}
\beta_{ti}, \beta_{s}= \boldsymbol{\Phi}_\mathrm{fit\_tissue} \left({D}_{\mathrm{recon}}  \right)
\label{fittissue}\quad,
\end{equation}
which enables us to keep the skull fixed while sampling perturbed coefficients to generate diverse facial variations. Figure~\ref{recon_res2} provides more reconstruction results of our method.

\subsection{Surgical Planning Prediction}
To make FSMM and TMM practical for real-world applications, we target clinical craniofacial surgery scenarios. Before surgery, clinicians typically edit the skull geometry based on the patient’s original skull mesh to formulate a surgical plan. A key step in this process is predicting the postoperative facial appearance corresponding to the modified skull, as it helps guide surgical decision-making and enhances doctor–patient communication.

\begin{figure}[t]
\centering
\includegraphics[width=0.98\columnwidth]{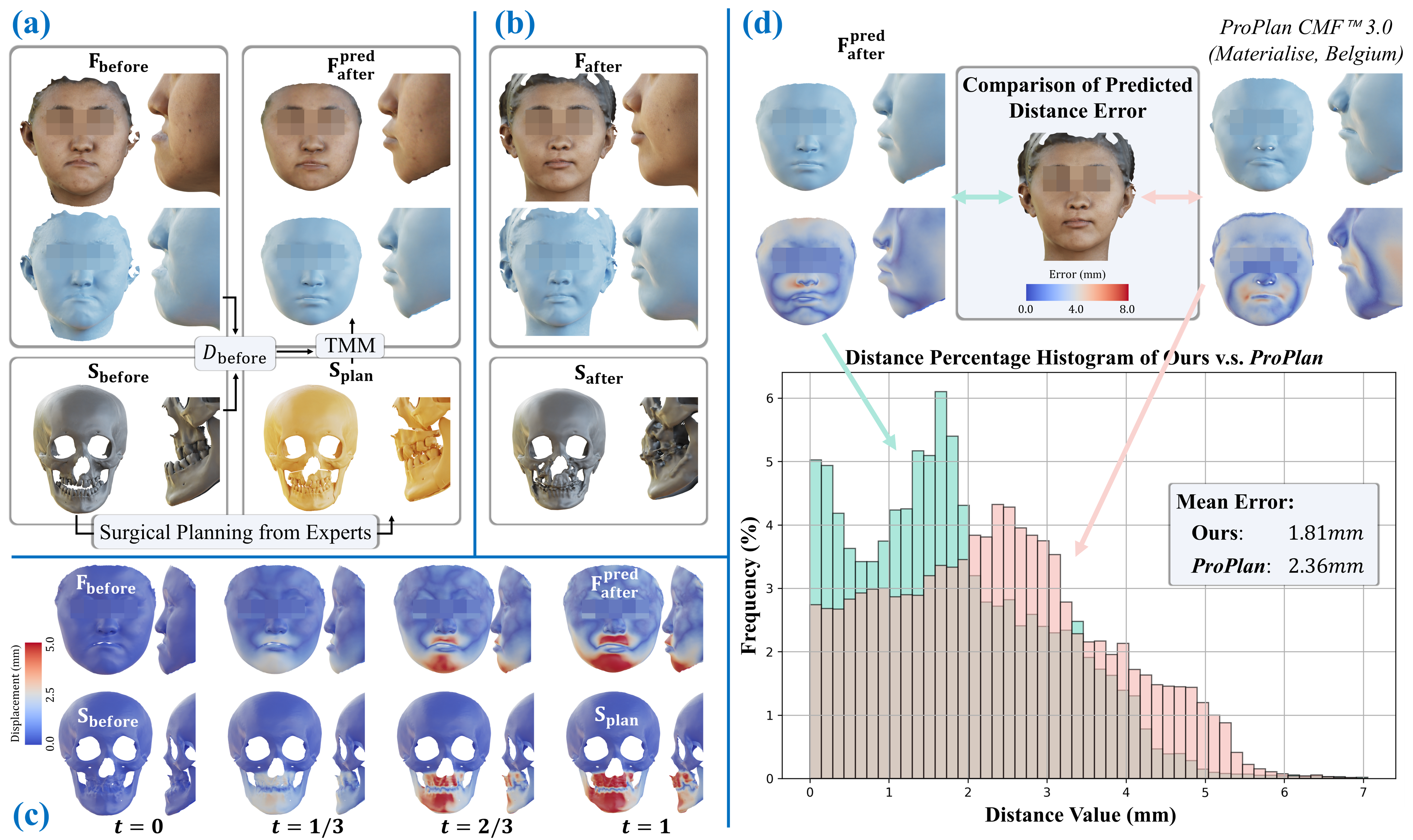}
\caption{Surgical planning prediction. (a) Our prediction. (b) Real postoperative outcome. (c) Interpolated deformation with visualized displacement. (d) Error comparison.}
\label{Surgical_Planning_Prediction}
\end{figure}

As shown in Figure~\ref{Surgical_Planning_Prediction}(a), given a skull modification plan \( \mathbf{S}_{\mathrm{plan}} \) specified by a clinician based on the current face \( \mathbf{F}_{\mathrm{before}} \) and skull \( \mathbf{S}_{\mathrm{before}} \), our method estimates the tissue thickness \( {D}_{\mathrm{before}} \) and uses it along with \( \mathbf{S}_{\mathrm{plan}} \) to predict the postoperative face \( \mathbf{F}_{\mathrm{after}}^{\mathrm{pred}} \). Figure~\ref{Surgical_Planning_Prediction}(b) shows the actual postoperative outcome, including the face \( \mathbf{F}_{\mathrm{after}} \) and the skull \( \mathbf{S}_{\mathrm{after}} \). Leveraging topological consistency, Figure~\ref{Surgical_Planning_Prediction}(c) further visualizes intermediate morphing results between the preoperative and planned meshes, with displacement heatmaps that can aid doctor–patient communication.

In clinical practice, software such as \textit{ProPlan CMF™ 3.0 (Materialise, Belgium)} is commonly used for postoperative prediction. As shown in Figure~\ref{Surgical_Planning_Prediction}(d), we compare the predicted distance errors from our method and \textit{ProPlan} against the real postoperative face \( \mathbf{F}_{\mathrm{after}} \). Our method yields lower prediction error ($1.81\,\mathrm{mm}$ \textit{vs.} $2.36\,\mathrm{mm}$).

\begin{table*}[t]
\centering
\small{
\setlength{\tabcolsep}{0.2mm} 
\begin{tabular}{c|cccccc|cccccc}
\toprule[1pt]
\multirow{3}{*}{Method} & \multicolumn{6}{c|}{\textit{Frontal-view Image as Input}}                                              & \multicolumn{6}{c}{\textit{Side-view Image as Input}}                                                  \\ \cline{2-13} 
                        & \multicolumn{3}{c|}{\textit{Face}}                        & \multicolumn{3}{c|}{\textit{Skull}}  & \multicolumn{3}{c|}{\textit{Face}}                        & \multicolumn{3}{c}{Skull}   \\ 
                        & NRMSE $\downarrow$   & Recall $\uparrow$  & \multicolumn{1}{c|}{Chamfer $\downarrow$} & NRMSE $\downarrow$   & Recall $\uparrow$  & Chamfer $\downarrow$ & NRMSE $\downarrow$   & Recall $\uparrow$  & \multicolumn{1}{c|}{Chamfer $\downarrow$} & NRMSE $\downarrow$   & Recall $\uparrow$  & Chamfer $\downarrow$ \\ \midrule[0.5pt]
SMIRK+FSMM                   & 3.169 & 0.803 & \multicolumn{1}{c|}{7.546} & 2.086 & 0.921 & 4.567 & 2.725 & 0.873 & \multicolumn{1}{c|}{6.781} & 1.884 & 0.942 & 4.160 \\
DECA+FSMM                    & 3.041 & 0.834 & \multicolumn{1}{c|}{7.645} & 2.022 & 0.932 & 4.512 & 2.941 & 0.830 & \multicolumn{1}{c|}{7.373} & 1.807 & 0.949 & 4.111 \\
3DDFA+FSMM                   & 4.251 & 0.702 & \multicolumn{1}{c|}{8.782} & 2.440 & 0.888 & 5.393 & 3.491 & 0.796 & \multicolumn{1}{c|}{7.196} & 2.250 & 0.908 & 5.054 \\
Deep3D+FSMM                  & 2.885 & 0.840 & \multicolumn{1}{c|}{5.890} & 1.943 & 0.935 & 4.295 & 2.496 & 0.897 & \multicolumn{1}{c|}{5.112} & 1.728 & 0.957 & 3.914 \\
Ours                    & \textbf{2.829} & \textbf{0.847} & \multicolumn{1}{c|}{\textbf{5.753}} & \textbf{1.933} & \textbf{0.935} & \textbf{4.283} & \textbf{2.420} & \textbf{0.911} & \multicolumn{1}{c|}{\textbf{4.963}} & \textbf{1.726} & \textbf{0.958} & \textbf{3.911} \\ \bottomrule[1pt]\end{tabular}
}
\caption{Quantitative comparison of 3D face–skull reconstruction from images. The best is highlighted in bold.}
\label{Inference-based}
\end{table*}

\begin{table}[t]
\centering
\small{
\setlength{\tabcolsep}{0.6mm} 
\begin{tabular}{c|cc|ccc}
\toprule[1pt]
Method                 & $n_{id}$ & $n_{exp}$ & NRMSE $\downarrow$  & Recall $\uparrow$ & Chamfer $\downarrow$ \\ \midrule[0.5pt]
\multirow{2}{*}{FLAME} & 300 & 0    & 0.520 & 0.910 & 7.539  \\
                       & 300 & 100  & 0.497 & 0.918 & 7.527   \\ \midrule[0.5pt]
\multirow{2}{*}{BFM}   & 199 & 0    & 0.766 & 0.726 & 1.881  \\
                       & 199 & 79$^\dagger$   & 0.616 & 0.814 & 1.625 \\ \midrule[0.5pt]
\multirow{4}{*}{Ours}  & 199 & 0    & 0.448 & 0.909 & 1.275  \\
                       & 300 & 0    & 0.421 & 0.923 & 1.230  \\
                       & 322 & 0    & 0.419 & 0.924 & 1.226  \\
                       & 322 & 79$^\dagger$   & \textbf{0.340} & \textbf{0.960} & \textbf{1.049}  \\ \bottomrule[1pt]
\end{tabular}
}
\caption{Quantitative comparison of the upper bounds of different face model representation capabilities.}
\begin{tablenotes}
\small
\item $^\dagger$ FaceWarehouse is used as the expression basis for BFM to enhance its representational capacity for fair comparison.
\end{tablenotes}
\label{optimize-based}
\end{table}

\begin{table}[t]
\centering
\small{
\setlength{\tabcolsep}{0.6mm} 
\begin{tabular}{cc|cc|cc}
\toprule[1pt]
\multicolumn{2}{c|}{$N_{q} = 1$ } & \multicolumn{2}{c|}{$N_{q} = 3$ } & \multicolumn{2}{c}{$N_{q} = 5$ }  \\ 
$N_{O}$  & NRMSE $\downarrow$  & $N_{O}$ & NRMSE $\downarrow$  & $N_{O}$ & NRMSE $\downarrow$  \\ \midrule[0.5pt]
$N_\mathrm{F}/100$             & 0.600 & $N_\mathrm{F}/100$             & 0.599 & $N_\mathrm{F}/100$             & 0.598 \\
$N_\mathrm{F}/10$               & 0.596 & $N_\mathrm{F}/10$               & 0.594 & $N_\mathrm{F}/10$              & 0.592 \\
$N_\mathrm{F}$                    & 0.594 & $N_\mathrm{F}$                 &\textbf{ 0.588} & $N_\mathrm{F}$                   & 0.590 \\ \bottomrule[1pt]
\end{tabular}
}
\caption{Ablation study for registration.}
\label{AblationStudytable}
\end{table}

\section{Experiments}

\subsection{Experimental Settings}
\myparagraph{Data.} We use 161 identities (322 samples with flipping) to build the FSMM and TMM. For the remaining samples, we randomly select 10 (20 with flipping) for ground-truth testing, covering both normal subjects and cases such as hemifacial microsomia, maxillary retrusion, mandibular protrusion, midface depression, and bimaxillary protrusion. For the comparison of image-based reconstruction accuracy, we render the test samples from multiple viewpoints to generate input for the evaluated methods. To train \( \boldsymbol{\Phi}_\mathrm{recon} \), 2D in-the-wild face images from publicly available datasets \cite{liu2015faceattributes, sagonas2013300, CelebAMask-HQ} with pose augmentation \cite{zhu2017face} are used, excluding those with facial expressions to better align with FSMM. The training set for \( \boldsymbol{\Phi}_\mathrm{recon} \) contains approximately $150K$ face images.

\myparagraph{Implementation Details.} We implement BFSM mainly based on PyTorch \cite{paszke2019pytorch}, Nvdiffrast \cite{Laine2020diffrast}, PyTorch3D \cite{ravi2020pytorch3d} and Kaolin \cite{KaolinLibrary}. The input image ${I_{face}}$ for \( \boldsymbol{\Phi}_\mathrm{recon} \) is resized into $224\cdot 224$. \(N_\mathrm{F} = 35709\). \(N_\mathrm{S} = 23208\).

\myparagraph{SOTA Methods for Comparison.} To evaluate the upper bound of BSFM in 3D face representation, we compare our model against state-of-the-art morphable models, including BFM \cite{blanz1999morphable}, and FLAME \cite{FLAME:SiggraphAsia2017}. FaceWarehouse \cite{cao2013facewarehouse} is used as the expression basis for BFM to enhance its representational capacity on certain deformity cases, enabling a fair comparison with BSFM. To evaluate the accuracy of paired reconstruction of the face and skull from images, we compare our method with state-of-the-art approaches \cite{SMIRK,DECA:Siggraph2021,wang20243d,deng2019accurate}. As existing methods reconstruct only facial geometry, we extend their outputs using FSMM to estimate the skull, enabling a comprehensive evaluation of face–skull reconstruction.

\myparagraph{Metrics.} Accuracy is measured using Normalized Root Mean Square Error (NRMSE) and Chamfer Distance. For Chamfer Distance, the GT is cropped to match the predicted region. Recall \cite{tatarchenko2019single, giebenhain2024mononphm} quantifies the percentage of GT points that have a nearby point in the reconstruction within a threshold.

\subsection{Comparison}
\label{subsec:QuantitativeComparison}
Table~\ref{Inference-based} presents the quantitative comparison of image-based reconstruction accuracy, where our method achieves the best overall performance. Note that skull reconstructions for compared methods are obtained by applying our FSMM to their respective face reconstruction outputs. Table~\ref{optimize-based} presents the upper bounds for different face model representations, highlighting the advantages of our method. These results are obtained by optimizing each model to minimize reconstruction errors with respect to the 3D GT mesh, thereby reflecting the best possible representation each model can achieve. As a result, the reported errors in Table~\ref{optimize-based} are lower than those in Table~\ref{Inference-based}. Distance errors are reported in millimeters.

\subsection{Ablation Study}
\label{AblationStudy}

Table~\ref{AblationStudytable} shows that increasing the number of ray origins (\(N_O\)) may improve registration accuracy by using denser control point correspondences, while setting the number of endpoints to \(N_q = 3\) yields the best overall performance.

\section{Conclusions}
We focus on paired face and skull modeling and reconstruction, introducing a large dataset and a novel dense ray matching registration method that ensures topological consistency across normal and pathological cases. We propose the 3D Bidirectional Face–Skull Morphable Model (BFSM) that enables accurate bidirectional inference between face and skull geometry and models tissue variability for diverse reconstructions. Experiments demonstrate that our BFSM outperforms existing state-of-the-art methods and shows promise for many applications.

{
    \small
    \bibliographystyle{ieeenat_fullname}
    \bibliography{skull}

\begin{thebibliography}{51}
\providecommand{\natexlab}[1]{#1}
\providecommand{\url}[1]{\texttt{#1}}
\expandafter\ifx\csname urlstyle\endcsname\relax
  \providecommand{\doi}[1]{doi: #1}\else
  \providecommand{\doi}{doi: \begingroup \urlstyle{rm}\Url}\fi

\bibitem[nic(2021)]{nicp_github}
Gpu accelerated non-rigid icp for surface registration.
\newblock \url{https://github.com/wuhaozhe/pytorch-nicp}, 2021.

\bibitem[Achenbach et~al.(2018)Achenbach, Brylka, Gietzen, Zum~Hebel, Sch{\"o}mer, Schulze, Botsch, and Schwanecke]{achenbach2018multilinear}
Jascha Achenbach, Robert Brylka, Thomas Gietzen, Katja Zum~Hebel, Elmar Sch{\"o}mer, Ralf Schulze, Mario Botsch, and Ulrich Schwanecke.
\newblock A multilinear model for bidirectional craniofacial reconstruction.
\newblock In \emph{VCBM@ MICCAI}, pages 67--76, 2018.

\bibitem[Amberg et~al.(2007)Amberg, Romdhani, and Vetter]{amberg2007optimal}
Brian Amberg, Sami Romdhani, and Thomas Vetter.
\newblock Optimal step nonrigid icp algorithms for surface registration.
\newblock In \emph{2007 IEEE conference on computer vision and pattern recognition}, pages 1--8. IEEE, 2007.

\bibitem[Blanz and Vetter(1999)]{blanz1999morphable}
Volker Blanz and Thomas Vetter.
\newblock A morphable model for the synthesis of 3d faces.
\newblock In \emph{Proceedings of the 26th annual conference on Computer graphics and interactive techniques}, pages 187--194, 1999.

\bibitem[Blanz and Vetter(2003)]{blanz2003face}
Volker Blanz and Thomas Vetter.
\newblock Face recognition based on fitting a 3d morphable model.
\newblock \emph{IEEE Transactions on pattern analysis and machine intelligence}, 25\penalty0 (9):\penalty0 1063--1074, 2003.

\bibitem[Cao et~al.(2013)Cao, Weng, Zhou, Tong, and Zhou]{cao2013facewarehouse}
Chen Cao, Yanlin Weng, Shun Zhou, Yiying Tong, and Kun Zhou.
\newblock Facewarehouse: A 3d facial expression database for visual computing.
\newblock \emph{IEEE Transactions on Visualization and Computer Graphics}, 20\penalty0 (3):\penalty0 413--425, 2013.

\bibitem[Chandran and Zoss(2024)]{chandran2024anatomically}
Prashanth Chandran and Gaspard Zoss.
\newblock Anatomically constrained implicit face models.
\newblock In \emph{Proceedings of the IEEE/CVF Conference on Computer Vision and Pattern Recognition}, pages 2220--2229, 2024.

\bibitem[Dai et~al.(2020)Dai, Pears, Smith, and Duncan]{dai2020statisticalLYHM}
Hang Dai, Nick Pears, William Smith, and Christian Duncan.
\newblock Statistical modeling of craniofacial shape and texture.
\newblock \emph{International Journal of Computer Vision}, 128:\penalty0 547--571, 2020.

\bibitem[Deng et~al.(2019)Deng, Yang, Xu, Chen, Jia, and Tong]{deng2019accurate}
Yu Deng, Jiaolong Yang, Sicheng Xu, Dong Chen, Yunde Jia, and Xin Tong.
\newblock Accurate 3d face reconstruction with weakly-supervised learning: From single image to image set.
\newblock In \emph{Proceedings of the IEEE/CVF conference on computer vision and pattern recognition workshops}, pages 0--0, 2019.

\bibitem[Feng et~al.(2021)Feng, Feng, Black, and Bolkart]{DECA:Siggraph2021}
Yao Feng, Haiwen Feng, Michael~J. Black, and Timo Bolkart.
\newblock Learning an animatable detailed {3D} face model from in-the-wild images.
\newblock 2021.

\bibitem[Foti et~al.(2022)Foti, Koo, Stoyanov, and Clarkson]{foti20223d}
Simone Foti, Bongjin Koo, Danail Stoyanov, and Matthew~J Clarkson.
\newblock 3d shape variational autoencoder latent disentanglement via mini-batch feature swapping for bodies and faces.
\newblock In \emph{Proceedings of the IEEE/CVF conference on computer vision and pattern recognition}, pages 18730--18739, 2022.

\bibitem[Fuji~Tsang et~al.(2022)Fuji~Tsang, Shugrina, Lafleche, Takikawa, Wang, Loop, Chen, Jatavallabhula, Smith, Rozantsev, Perel, Shen, Gao, Fidler, State, Gorski, Xiang, Li, Li, and Lebaredian]{KaolinLibrary}
Clement Fuji~Tsang, Maria Shugrina, Jean~Francois Lafleche, Towaki Takikawa, Jiehan Wang, Charles Loop, Wenzheng Chen, Krishna~Murthy Jatavallabhula, Edward Smith, Artem Rozantsev, Or Perel, Tianchang Shen, Jun Gao, Sanja Fidler, Gavriel State, Jason Gorski, Tommy Xiang, Jianing Li, Michael Li, and Rev Lebaredian.
\newblock Kaolin: A pytorch library for accelerating 3d deep learning research.
\newblock \url{https://github.com/NVIDIAGameWorks/kaolin}, 2022.

\bibitem[Giebenhain et~al.(2023)Giebenhain, Kirschstein, Georgopoulos, R{\"{u}}nz, Agapito, and Nie{\ss}ner]{giebenhain2023nphm}
Simon Giebenhain, Tobias Kirschstein, Markos Georgopoulos, Martin R{\"{u}}nz, Lourdes Agapito, and Matthias Nie{\ss}ner.
\newblock Learning neural parametric head models.
\newblock In \emph{Proc. IEEE Conf. on Computer Vision and Pattern Recognition (CVPR)}, 2023.

\bibitem[Giebenhain et~al.(2024)Giebenhain, Kirschstein, Georgopoulos, R{\"{u}}nz, Agapito, and Nie{\ss}ner]{giebenhain2024mononphm}
Simon Giebenhain, Tobias Kirschstein, Markos Georgopoulos, Martin R{\"{u}}nz, Lourdes Agapito, and Matthias Nie{\ss}ner.
\newblock Mononphm: Dynamic head reconstruction from monoculuar videos.
\newblock In \emph{Proc. IEEE Conf. on Computer Vision and Pattern Recognition (CVPR)}, 2024.

\bibitem[Jolliffe and Cadima(2016)]{jolliffe2016principal}
Ian~T Jolliffe and Jorge Cadima.
\newblock Principal component analysis: a review and recent developments.
\newblock \emph{Philosophical transactions of the royal society A: Mathematical, Physical and Engineering Sciences}, 374\penalty0 (2065):\penalty0 20150202, 2016.

\bibitem[Keller et~al.(2022)Keller, Zuffi, Black, and Pujades]{keller2022osso}
Marilyn Keller, Silvia Zuffi, Michael~J Black, and Sergi Pujades.
\newblock Osso: Obtaining skeletal shape from outside.
\newblock In \emph{Proceedings of the IEEE/CVF conference on computer vision and pattern recognition}, pages 20492--20501, 2022.

\bibitem[Keller et~al.(2023)Keller, Werling, Shin, Delp, Pujades, Liu, and Black]{keller2023skin}
Marilyn Keller, Keenon Werling, Soyong Shin, Scott Delp, Sergi Pujades, C~Karen Liu, and Michael~J Black.
\newblock From skin to skeleton: Towards biomechanically accurate 3d digital humans.
\newblock \emph{ACM Transactions on Graphics (TOG)}, 42\penalty0 (6):\penalty0 1--12, 2023.

\bibitem[Keller et~al.(2024)Keller, Arora, Dakri, Chandhok, Machann, Fritsche, Black, and Pujades]{keller2024hit}
Marilyn Keller, Vaibhav Arora, Abdelmouttaleb Dakri, Shivam Chandhok, J{\"u}rgen Machann, Andreas Fritsche, Michael~J Black, and Sergi Pujades.
\newblock Hit: Estimating internal human implicit tissues from the body surface.
\newblock In \emph{Proceedings of the IEEE/CVF Conference on Computer Vision and Pattern Recognition}, pages 3480--3490, 2024.

\bibitem[Laine et~al.(2020)Laine, Hellsten, Karras, Seol, Lehtinen, and Aila]{Laine2020diffrast}
Samuli Laine, Janne Hellsten, Tero Karras, Yeongho Seol, Jaakko Lehtinen, and Timo Aila.
\newblock Modular primitives for high-performance differentiable rendering.
\newblock \emph{ACM Transactions on Graphics}, 39\penalty0 (6), 2020.

\bibitem[Lee et~al.(2020)Lee, Liu, Wu, and Luo]{CelebAMask-HQ}
Cheng-Han Lee, Ziwei Liu, Lingyun Wu, and Ping Luo.
\newblock Maskgan: Towards diverse and interactive facial image manipulation.
\newblock In \emph{IEEE Conference on Computer Vision and Pattern Recognition (CVPR)}, 2020.

\bibitem[Li et~al.(2017)Li, Bolkart, Black, Li, and Romero]{FLAME:SiggraphAsia2017}
Tianye Li, Timo Bolkart, Michael.~J. Black, Hao Li, and Javier Romero.
\newblock Learning a model of facial shape and expression from {4D} scans.
\newblock \emph{ACM Transactions on Graphics, (Proc. SIGGRAPH Asia)}, 36\penalty0 (6):\penalty0 194:1--194:17, 2017.

\bibitem[Liang et~al.(2024)Liang, Zhang, Zhao, Wang, and Li]{10791299SkulltoFace}
Yongqing Liang, Congyi Zhang, Junli Zhao, Wenping Wang, and Xin Li.
\newblock Skull-to-face: Anatomy-guided 3d facial reconstruction and editing.
\newblock \emph{IEEE Transactions on Visualization and Computer Graphics}, pages 1--13, 2024.

\bibitem[Liu et~al.(2015)Liu, Luo, Wang, and Tang]{liu2015faceattributes}
Ziwei Liu, Ping Luo, Xiaogang Wang, and Xiaoou Tang.
\newblock Deep learning face attributes in the wild.
\newblock In \emph{Proceedings of International Conference on Computer Vision (ICCV)}, 2015.

\bibitem[Loper et~al.(2015)Loper, Mahmood, Romero, Pons-Moll, and Black]{SMPL:2015}
Matthew Loper, Naureen Mahmood, Javier Romero, Gerard Pons-Moll, and Michael~J. Black.
\newblock {SMPL}: A skinned multi-person linear model.
\newblock \emph{ACM Trans. Graphics (Proc. SIGGRAPH Asia)}, 34\penalty0 (6):\penalty0 248:1--248:16, 2015.

\bibitem[L{\"u}thi et~al.(2009)L{\"u}thi, Lerch, Albrecht, Krol, and Vetter]{luthi2009hierarchical}
Marcel L{\"u}thi, Anita Lerch, Thomas Albrecht, Zdzislaw Krol, and Thomas Vetter.
\newblock A hierarchical, multi-resolution approach for model-based skull-segmentation in mri volumes.
\newblock In \emph{Conference Proceedings D}, 2009.

\bibitem[Madsen et~al.(2018)Madsen, L{\"u}thi, Schneider, and Vetter]{madsen2018probabilistic}
Dennis Madsen, Marcel L{\"u}thi, Andreas Schneider, and Thomas Vetter.
\newblock Probabilistic joint face-skull modelling for facial reconstruction.
\newblock In \emph{Proceedings of the IEEE conference on computer vision and pattern recognition}, pages 5295--5303, 2018.

\bibitem[Mildenhall et~al.(2020)Mildenhall, Srinivasan, Tancik, Barron, Ramamoorthi, and Ng]{mildenhall2020nerf}
Ben Mildenhall, Pratul~P. Srinivasan, Matthew Tancik, Jonathan~T. Barron, Ravi Ramamoorthi, and Ren Ng.
\newblock Nerf: Representing scenes as neural radiance fields for view synthesis.
\newblock In \emph{ECCV}, 2020.

\bibitem[Paszke et~al.(2019)Paszke, Gross, Massa, Lerer, Bradbury, Chanan, Killeen, Lin, Gimelshein, Antiga, et~al.]{paszke2019pytorch}
Adam Paszke, Sam Gross, Francisco Massa, Adam Lerer, James Bradbury, Gregory Chanan, Trevor Killeen, Zeming Lin, Natalia Gimelshein, Luca Antiga, et~al.
\newblock Pytorch: An imperative style, high-performance deep learning library.
\newblock \emph{Advances in neural information processing systems}, 32, 2019.

\bibitem[Paysan et~al.(2009)Paysan, L{\"u}thi, Albrecht, Lerch, Amberg, Santini, and Vetter]{paysan2009face}
Pascal Paysan, Marcel L{\"u}thi, Thomas Albrecht, Anita Lerch, Brian Amberg, Francesco Santini, and Thomas Vetter.
\newblock Face reconstruction from skull shapes and physical attributes.
\newblock In \emph{Pattern Recognition: 31st DAGM Symposium, Jena, Germany, September 9-11, 2009. Proceedings 31}, pages 232--241. Springer, 2009.

\bibitem[Porto et~al.(2021)Porto, Rolfe, and Maga]{porto2021alpaca}
Arthur Porto, Sara Rolfe, and A~Murat Maga.
\newblock Alpaca: A fast and accurate computer vision approach for automated landmarking of three-dimensional biological structures.
\newblock \emph{Methods in Ecology and Evolution}, 12\penalty0 (11):\penalty0 2129--2144, 2021.

\bibitem[Potamias et~al.(2024)Potamias, Tarasiou, Ploumpis, and Zafeiriou]{potamias2024shapefusion}
Rolandos~Alexandros Potamias, Michail Tarasiou, Stylianos Ploumpis, and Stefanos Zafeiriou.
\newblock Shapefusion: A 3d diffusion model for localized shape editing.
\newblock In \emph{European Conference on Computer Vision}, pages 72--89. Springer, 2024.

\bibitem[Ramamoorthi and Hanrahan(2001)]{ramamoorthi2001efficient}
Ravi Ramamoorthi and Pat Hanrahan.
\newblock An efficient representation for irradiance environment maps.
\newblock In \emph{Proceedings of the 28th annual conference on Computer graphics and interactive techniques}, pages 497--500, 2001.

\bibitem[Ravi et~al.(2020)Ravi, Reizenstein, Novotny, Gordon, Lo, Johnson, and Gkioxari]{ravi2020pytorch3d}
Nikhila Ravi, Jeremy Reizenstein, David Novotny, Taylor Gordon, Wan-Yen Lo, Justin Johnson, and Georgia Gkioxari.
\newblock Accelerating 3d deep learning with pytorch3d.
\newblock \emph{arXiv:2007.08501}, 2020.

\bibitem[Reddi et~al.(2019)Reddi, Kale, and Kumar]{reddi2019convergence}
Sashank~J Reddi, Satyen Kale, and Sanjiv Kumar.
\newblock On the convergence of adam and beyond.
\newblock \emph{arXiv preprint arXiv:1904.09237}, 2019.

\bibitem[Retsinas et~al.(2024)Retsinas, Filntisis, Danecek, Abrevaya, Roussos, Bolkart, and Maragos]{SMIRK}
George Retsinas, Panagiotis~P Filntisis, Radek Danecek, Victoria~F Abrevaya, Anastasios Roussos, Timo Bolkart, and Petros Maragos.
\newblock 3d facial expressions through analysis-by-neural-synthesis.
\newblock In \emph{Proceedings of the IEEE/CVF Conference on Computer Vision and Pattern Recognition}, pages 2490--2501, 2024.

\bibitem[Sagonas et~al.(2013)Sagonas, Tzimiropoulos, Zafeiriou, and Pantic]{sagonas2013300}
Christos Sagonas, Georgios Tzimiropoulos, Stefanos Zafeiriou, and Maja Pantic.
\newblock 300 faces in-the-wild challenge: The first facial landmark localization challenge.
\newblock In \emph{Proceedings of the IEEE international conference on computer vision workshops}, pages 397--403, 2013.

\bibitem[Sirovich and Kirby(1987)]{Sirovich:87}
L. Sirovich and M. Kirby.
\newblock Low-dimensional procedure for the characterization of human faces.
\newblock \emph{J. Opt. Soc. Am. A}, 4\penalty0 (3):\penalty0 519--524, 1987.

\bibitem[Tarasiou et~al.(2024)Tarasiou, Potamias, O'Sullivan, Ploumpis, and Zafeiriou]{tarasiou2024locally}
Michail Tarasiou, Rolandos~Alexandros Potamias, Eimear O'Sullivan, Stylianos Ploumpis, and Stefanos Zafeiriou.
\newblock Locally adaptive neural 3d morphable models, 2024.

\bibitem[Tatarchenko et~al.(2019)Tatarchenko, Richter, Ranftl, Li, Koltun, and Brox]{tatarchenko2019single}
Maxim Tatarchenko, Stephan~R Richter, Ren{\'e} Ranftl, Zhuwen Li, Vladlen Koltun, and Thomas Brox.
\newblock What do single-view 3d reconstruction networks learn?
\newblock In \emph{Proceedings of the IEEE/CVF conference on computer vision and pattern recognition}, pages 3405--3414, 2019.

\bibitem[Wang et~al.(2022)Wang, Chen, Yu, Ma, Li, and Liu]{wang2022faceverse}
Lizhen Wang, Zhiyuan Chen, Tao Yu, Chenguang Ma, Liang Li, and Yebin Liu.
\newblock Faceverse: a fine-grained and detail-controllable 3d face morphable model from a hybrid dataset.
\newblock In \emph{Proceedings of the IEEE/CVF Conference on Computer Vision and Pattern Recognition}, pages 20333--20342, 2022.

\bibitem[Wang et~al.(2021)Wang, Liu, Liu, Theobalt, Komura, and Wang]{wang2021neus}
Peng Wang, Lingjie Liu, Yuan Liu, Christian Theobalt, Taku Komura, and Wenping Wang.
\newblock Neus: Learning neural implicit surfaces by volume rendering for multi-view reconstruction.
\newblock \emph{arXiv preprint arXiv:2106.10689}, 2021.

\bibitem[Wang et~al.(2024{\natexlab{a}})Wang, Zhu, Yu, Zhang, and Lei]{wang2024s2td}
Zidu Wang, Xiangyu Zhu, Jiang Yu, Tianshuo Zhang, and Zhen Lei.
\newblock S2td-face: Reconstruct a detailed 3d face with controllable texture from a single sketch.
\newblock In \emph{Proceedings of the 32nd ACM International Conference on Multimedia}, pages 6453--6462, 2024{\natexlab{a}}.

\bibitem[Wang et~al.(2024{\natexlab{b}})Wang, Zhu, Zhang, Wang, and Lei]{wang20243d}
Zidu Wang, Xiangyu Zhu, Tianshuo Zhang, Baiqin Wang, and Zhen Lei.
\newblock 3d face reconstruction with the geometric guidance of facial part segmentation.
\newblock In \emph{Proceedings of the IEEE/CVF Conference on Computer Vision and Pattern Recognition}, pages 1672--1682, 2024{\natexlab{b}}.

\bibitem[Wang et~al.(2025)Wang, Zhao, Xu, Zhu, and Lei]{wang2025srmhair}
Zidu Wang, Jiankuo Zhao, Miao Xu, Xiangyu Zhu, and Zhen Lei.
\newblock Srm-hair: Single image head mesh reconstruction via 3d morphable hair, 2025.

\bibitem[Wood et~al.(2021)Wood, Baltru{\v{s}}aitis, Hewitt, Dziadzio, Cashman, and Shotton]{wood2021fake}
Erroll Wood, Tadas Baltru{\v{s}}aitis, Charlie Hewitt, Sebastian Dziadzio, Thomas~J Cashman, and Jamie Shotton.
\newblock Fake it till you make it: face analysis in the wild using synthetic data alone.
\newblock In \emph{Proceedings of the IEEE/CVF international conference on computer vision}, pages 3681--3691, 2021.

\bibitem[Yang et~al.(2020)Yang, Zhu, Wang, Huang, Shen, Yang, and Cao]{yang2020facescape}
Haotian Yang, Hao Zhu, Yanru Wang, Mingkai Huang, Qiu Shen, Ruigang Yang, and Xun Cao.
\newblock Facescape: A large-scale high quality 3d face dataset and detailed riggable 3d face prediction.
\newblock In \emph{IEEE/CVF Conference on Computer Vision and Pattern Recognition (CVPR)}, 2020.

\bibitem[Yang et~al.(2022)Yang, Kim, Zoss, G{\"o}zc{\"u}, Gross, and Solenthaler]{yang2022implicit}
Lingchen Yang, Byungsoo Kim, Gaspard Zoss, Baran G{\"o}zc{\"u}, Markus Gross, and Barbara Solenthaler.
\newblock Implicit neural representation for physics-driven actuated soft bodies.
\newblock \emph{ACM Transactions on Graphics (TOG)}, 41\penalty0 (4):\penalty0 1--10, 2022.

\bibitem[Yang et~al.(2023)Yang, Zoss, Chandran, et~al.]{yang2023implicit}
L. Yang, G. Zoss, P. Chandran, et~al.
\newblock An implicit physical face model driven by expression and style.
\newblock In \emph{SIGGRAPH Asia 2023 Conference Papers}, pages 1--12, 2023.

\bibitem[Yang et~al.(2024)Yang, Zoss, Chandran, Gross, Solenthaler, Sifakis, and Bradley]{yang2024learning}
Lingchen Yang, Gaspard Zoss, Prashanth Chandran, Markus Gross, Barbara Solenthaler, Eftychios Sifakis, and Derek Bradley.
\newblock Learning a generalized physical face model from data.
\newblock \emph{arXiv preprint arXiv:2402.19477}, 2024.

\bibitem[Zhu et~al.(2017)Zhu, Liu, Lei, and Li]{zhu2017face}
Xiangyu Zhu, Xiaoming Liu, Zhen Lei, and Stan~Z Li.
\newblock Face alignment in full pose range: A 3d total solution.
\newblock \emph{IEEE transactions on pattern analysis and machine intelligence}, 41\penalty0 (1):\penalty0 78--92, 2017.

\bibitem[Zoss et~al.(2019)Zoss, Beeler, Gross, et~al.]{zoss2019accurate}
G. Zoss, T. Beeler, M. Gross, et~al.
\newblock Accurate markerless jaw tracking for facial performance capture.
\newblock \emph{ACM Transactions on Graphics}, 38\penalty0 (4):\penalty0 1--8, 2019.

\end{thebibliography}
}

\end{document}